\documentclass[10pt,journal]{IEEEtran}

\usepackage{epsfig}
\usepackage{graphicx}
\usepackage{amsmath}
\usepackage{amssymb}
\usepackage{amsthm}

\usepackage{comment}
\usepackage{enumitem}
\usepackage{pifont}
\usepackage{url}
\usepackage{booktabs}
\usepackage{makecell}
\usepackage{multirow}
\usepackage{caption}
\usepackage{gensymb}
\usepackage[colorlinks,linkcolor=black,anchorcolor=black,citecolor=black,urlcolor=black]{hyperref}
\usepackage{algorithm}
\usepackage{rotating}
\usepackage[caption=false]{subfig}

\usepackage{bbding}
\usepackage{tabularx}

\newtheorem{theorem}{Theorem}

\usepackage{algpseudocode}%
\usepackage[table]{xcolor}

\ifCLASSOPTIONcompsoc
  \usepackage[nocompress]{cite}
\else
  \usepackage{cite}
\fi

\ifCLASSINFOpdf
\else
\fi

\begin{document}

\title{Generative Classifier for Domain Generalization}

\author{Shaocong Long, Qianyu Zhou, Dexuan Zhang, Tatsuya Harada
\IEEEcompsocitemizethanks{
Manuscript received XX XX 2026; revised XX XX 2026 and XX XX 2026; accepted XX XXX 2026. Date of publication XX 2026; date of current version 14 XX 2026. (\textit{Corresponding author: Qianyu Zhou}.)

\IEEEcompsocthanksitem S. Long are with the Department of Computer Science and Engineering, Shanghai Jiao Tong University, Shanghai 200240, China (e-mail: longshaocong@sjtu.edu.cn). \protect
\IEEEcompsocthanksitem Q. Zhou, D. Zhang, T. Harada are with Research Center for Advanced Science and Technology, The University of Tokyo, Japan, (e-mail: \{q-zhou, dexuan.zhang, harada\}@mi.t.u-tokyo.ac.jp)). \protect

This work is supported in part by the National Natural Science Foundation of China (NSFC) Young Scientists Fund (Category C)  (Grant No. 62502178), in part by the Open Project Program of the State Key Laboratory of CAD\&CG (Grant No. A2610), Zhejiang University, in part by JSPS KAKENHI Grant Number JP26K21298, in part by JST Moonshot R\&D Grant Number JPMJPS2011, in part by CREST Grant Number JPMJCR2015 and Basic Research Grant (Super AI) of Institute for AI and Beyond of the University of Tokyo.
}
}


\maketitle

\begin{abstract}
Domain generalization (DG) aims to improve the generalizability of computer vision models toward distribution shifts. 
The mainstream  DG methods predominantly focus on learning domain invariance across domains, however, such methods overlook the untapped potential inherent in domain-specific information. While the prevailing practice of discriminative linear classifier has been tailored to domain-invariant features, it struggles when confronted with diverse domain-specific information, \emph{e.g.,} intra-class shifts, that exhibits multi-modality. 
To address these issues, we explore the theoretical implications of relying on domain-invariant features, revealing the crucial role of domain-specific information in mitigating the target risk for DG. 
Drawing from these insights, we propose Generative Classifier-driven Domain Generalization (GCDG), introducing a generative paradigm for the DG classifier based on Gaussian Mixture Models (GMMs) for each class across domains. 
GCDG consists of three key modules: Heterogeneity Learning Classifier~(HLC), Spurious Correlation Blocking~(SCB), and Diverse Component Balancing~(DCB).
Concretely, HLC attempts to model the feature distributions and thereby capture valuable domain-specific information via GMMs.
SCB identifies the neural units containing spurious correlations and perturbs them, mitigating the risk of HLC learning irrelevant spurious patterns. 
Meanwhile, DCB ensures a balanced contribution of components within HLC, preventing the underestimation or neglect of critical components.
In this way, GCDG excels in capturing the nuances of domain-specific information characterized by diverse distributions. GCDG demonstrates the potential to reduce the target risk and encourage flat minima, improving the model's generalizability. 
Extensive experiments show GCDG's comparable performance on five DG benchmarks and one face anti-spoofing dataset, seamlessly integrating into existing DG methods with consistent improvements. 
Code will be available at \url{https://github.com/longshaocong/GCDG}.

\end{abstract}


\IEEEpeerreviewmaketitle
\section{Introduction}

\IEEEPARstart{L}{earning} a better visual representation~\cite{he2016deep,dosovitskiy2020image} has been widely explored with the fast development of deep learning.
Nevertheless, the persistent challenge of distribution shifts~\cite{zhou2023instance, meng2022attention, choi2021robustnet, zhou2021domain_tip} in real-world scenarios poses a substantial barrier to the deployment of trained models that assume an assumption of independent and identical distributions. 
For instance, printed photographs or replayed videos can deceive face recognition systems trained in environments that do not include such variations\cite{zhou2024test, yu2020fas, yu2020face}, creating potential security vulnerabilities in the system. 
To this end, enhancing models' generalization has become an urgent requirement. 
Domain generalization~(DG)~\cite{gulrajani2020search, wang2022generalizing, zhou2022domain, zhao2023style, li2023sparse,jiang2024dgpic} is one of the effective ways to alleviate the adverse effects of distribution shifts across domains, aiming to empower models with the capacity to discern and capture genuine patterns across diverse scenarios without access to any data in unseen target domains.

The mainstream approaches tend to excavate the \textit{domain-invariant features} across domains while suppressing domain-specific variations. 
Subsequently, a linear probe is fed with these features for classification. 
It has emerged as a prevalent paradigm in DG, encompassing various techniques, \emph{e.g.,} risk minimization~\cite{arjovsky2019invariant, long2025domain}, domain adversarial training~\cite{ganin2016domainadversarial, li2018deep, zhao2020domain, zhou2020deep}, contrastive learning~\cite{yao2022pcl, kim2021selfreg, cha2022domain, huang2023sentence}, feature disentanglement~\cite{zhang2022towards, wang2022domain_fre}, gradient invariance learning~\cite{shi2022gradient, mansilla2021domain, song2023gradca, rame2022fishr}. 
Despite the gratifying progress, enforcing strict domain invariance may lead to complete ignorance of domain-specific information that could aid the generalization~\cite{bui2021exploiting}, especially in scenarios involving complex samples or outliers~\cite{mahajan2021domain, yao2022pcl, lv2023improving}, leading to sub-optimal performance in unseen domains.

\begin{figure}
  \centering
  \includegraphics[width=\linewidth]{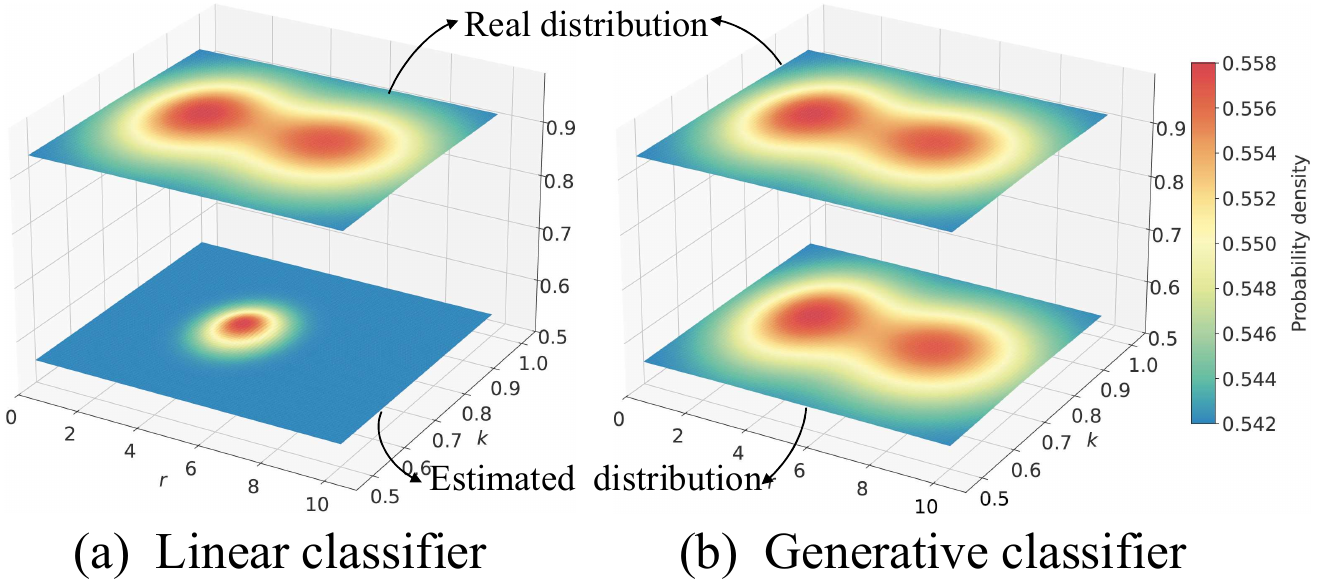}
  \caption{Comparison of modeling a class between the discriminative linear classifier and the proposed generative classifier in DG. (a) The prevailing linear classifier in DG operates under the assumption of unimodal distribution, encountering substantial challenges when confronted with domain-specific data that exhibits multi-modality. (b) In this paper, we introduce a novel generative classifier to capture the underlying multi-modal distribution present in domain-specific data.}
  \label{blue_print}
  \vspace{-5mm}
\end{figure}

To overcome the challenge posed by complex samples or outliers, specific domain-invariant methods are proposed, such as MatchDG~\cite{mahajan2021domain} and PCL~\cite{yao2022pcl}. These methods perform alignment within subdatasets, rather than aligning all data within a class across domains. 
However, these methods underscore the significance of differentiating sub-datasets within a class across domains to enhance generalization performance, implicitly suggesting that domain-invariant features may be detrimental to DG.

Another line of DG methods proposes to leverage the domain-specific features as the complementary information of the domain-invariant ones to improve the generalization capabilities. 
Such approaches focus on learning various domain experts~\cite{chattopadhyay2020learning, bui2021exploiting, zhou2022adaptive} or feature disentanglement\cite{zhang2022principled, wang2022variational} to first extract domain-specific characteristics and then complement the invariant ones for generalization. 
While the above methods have recognized the importance of domain-specific information in promoting generalization performance, they have overlooked a crucial issue: the common practice of utilizing a linear probe after the feature extractor could harm generalization performance.

The prevailing linear probe functions as a discriminative classifier in previous DG methods from a probabilistic perspective, proficiently delineating decision boundaries between classes.
However, such a discriminative linear classifier is inherently tailored for domain-invariant features and operates under a common assumption of unimodal distribution for each class. Thus, as shown in Fig.~\ref{blue_print}(a), it lacks the capability to harness valuable domain-specific information and may lead to sub-optimal performance when confronted with domain-specific information, \emph{i.e.,} distribution of each class exhibiting multi-modality across domains, which is a common occurrence in real-world scenarios. 
Moreover, as discriminative modeling, it inherently focuses on learning decision boundaries between classes while overlooking diverse feature distributions, making it difficult to fully capture the desirable domain-specific characteristics.
As such, the presence of domain-specific information calls for a more sophisticated approach than solely relying on a linear classifier to promote generalizability.

To effectively leverage domain-specific information, we propose exploring a generative paradigm as a potential choice for the DG classifier. 
Unlike discriminative learning, generative modeling captures the underlying data distribution, allowing a more comprehensive representation of domain-specific features. 
Some DG methods \cite{murkute2021domain, wang2022variational} have incorporated generative models (\emph{e.g.}, VAE). However, these methods primarily use them as auxiliary modules to impose additional constraints, such as image reconstruction and generation. As a result, these approaches remain focused on enforcing domain invariance rather than utilizing the generative paradigm to model feature distributions directly.
In contrast, our goal of introducing the generative paradigm is to capture domain-specific information by modeling the feature distribution, thereby unleashing the potential of domain-specific information.

In this work, we propose a new method, namely Generative Classifier-driven Domain Generalization (GCDG), a generative paradigm for DG classifier that replaces the discriminative classifier. By modeling the underlying diverse feature distributions, GCDG effectively leverages domain-specific information to lower the upper bound of target risk and thereby enhance generalizability.
GCDG comprises three key modules: Heterogeneity Learning Classifier~(HLC), Spurious Correlation Blocking~(SCB), and Diverse Component Balancing~(DCB).
The key to GCDG lies in HLC, which is a generative classifier and leverages Gaussian Mixture Models (GMM) for each class across domains, being able to model multi-modal distributions and represent a broader range of data patterns. 
While SCB aims to perturb harmful spurious correlations, preventing HLC from capturing them. 
Additionally, DCB constrains the uniform distribution of mixing coefficients in GMM, avoiding the underestimation or ignorance of components in HLC.

Our GCDG excels in capturing the nuances of domain-specific information characterized by diverse distributions, and has three superiorities: 
Firstly, unlike a linear probe that assumes unimodal distribution for each class, HLC's mixture components can effectively model the multi-modality in real-world scenarios, thus endowing GCDG with greater tolerance to intra-class variances. 
Secondly, the generative classifier can relax the feature alignment process by accommodating diverse features under identical one-hot labels, thereby promoting flat minima. 
Thirdly, when perfect matching is not required, the original information remains less compressed, leading to a lower upper bound of target risk. 
In summary, our main contributions are three-fold:
\vspace{-1mm}

\begin{itemize}
    \item We embark on a theoretical exploration of the implications arising from an increased upper bound of the target risk due to the reliance on domain-invariant features. We shed
    light on the crucial role that domain-specific information plays in reducing the target risk for DG. To the best of our knowledge, this is the first work studying insufficiency of the prevalent linear classifier in DG.

    \item We propose GCDG, a generative paradigm for DG classifier, comparing three key modules: HLC, SCB, and DCB. Concretely, HLC replaces the prevalent linear probe with the presented generative classifier, enabling the model to effectively capture diverse distributions across domains. SCB prevents HLC from capturing spurious correlations. Additionally, DCB ensures balanced contributions of components in HLC. 
    Consequently, GCDG encourages the reduction in the upper bound of target risk and the promotion of flat minima.
    
    \item Extensive experiments on five DG benchmarks and one face anti-spoofing benchmark demonstrate the effectiveness of the proposed GCDG against state-of-the-art competitors. Notably, GCDG could be seamlessly integrated with existing DG methods as a plug-and-play module with consistent performance improvements. 
\end{itemize}

\section{Related Work}

\noindent\textbf{DG Methods via Learning Domain-invariance.} The mainstream DG approaches aim to extract domain-invariant features while suppressing domain-specific information across domains. 
The intuitive approach for DG is to minimize the empirical source risks~\cite{vapnik1999overview, arjovsky2019invariant, lv2022causality, lin2022bayesian, michalkiewicz2023domain}. 
Domain adversarial training~\cite{ganin2016domainadversarial, zhao2020domain, zhou2020deep, long2024rethinking} seeks to align source distributions across domains, thereby acquiring common features. 
Contrastive learning~\cite{yao2022pcl, huang2023sentence, wang2022contrastive, qi2022class, huang2023idag, long2025domain} is another effective way to constrain the model to avoid learning spurious correlations.
Another key avenue is data augmentation~\cite{zhou2021domain, zhou2024mixstyle, xu2021fourier, zhao2022style, zhou2020learning, li2026causal, zhu2026causality}, which exposes the model to data of diverse styles.
Disentanglement~\cite{zhang2022towards, dai2023moderately, wang2022domain_fre, hu2023dandelionnet, zhu2026cfsm, cheng2026prompt} attempts to separate features into domain-invariant and domain-specific components, helping the model focus on domain invariance.
Methods based on gradient invariance~\cite{shi2022gradient, mansilla2021domain, song2023gradca} enforce domain invariance by imposing gradient constraints.
Frequency filtering techniques~\cite{guo2023aloft, lin2023deep} remove domain-specific frequency components, facilitating the learning of domain-invariant features.
Additionally, recent works explore the role of network architectures~\cite{li2023sparse, long2024dgmamba, guo2024seta} in improving generalization performance.  
Most DG methods in face anti-spoofing (FAS)~\cite{shao2019multi, hu2024rethinking, liu2023towards} also focus on learning domain invariance to enhance the security of face recognition systems.
Despite great progress, enforcing strict domain invariance may lead to complete ignorance of domain-specific features that could aid the generalization~\cite{bui2021exploiting}, leading to sub-optimal performance.

\noindent\textbf{DG Methods via Learning Domain-specificity.}
To address the above issue, certain studies propose to leverage the unique domain-specific characteristics as rich complementary information of the domain-invariant ones to enhance the generalization. Specifically, in the field of person re-identification (ReID) and FAS, recent works~\cite{chattopadhyay2020learning, zhou2022adaptive, dai2021generalizable} study various domain experts to learn discriminative domain-specific features as the complement of the domain-invariant ones, establishing the link between the seen domains and unseen domains and further improving the generalization. Another work~\cite{wang2022domain} introduces a feature disentanglement and information interaction mechanism to ensure the effective collaboration of domain-invariant and domain-specific information.
In test-time adaptation, DRM~\cite{zhang2023domain} explores and ensembles various domain-specific classifiers to minimize the adaptivity gap based on the target samples.
Besides, DMG~\cite{chattopadhyay2020learning} studies domain-specific masks and averages the predictions obtained by applying various masks, aiming to achieve a balance between domain-invariant and domain-specific features.

\noindent\textbf{Generative Classifiers.}
Methods across various fields have demonstrated the advantages of generative classifiers. 
In the realm of adversarial attacks, Deep Bayes~\cite{li2019generative} models the conditional distribution of inputs using a latent variable model, which helps verify the ``off-manifold" conjecture. 
\cite{zheng2023revisiting} show that naive Bayes leads to faster convergence than discriminative classifiers in pre-trained deep models. 
In semantic segmentation, GMMseg\cite{liang2022gmmseg} utilizes generative classifiers to capture class-conditional densities, combining the strengths of both generative and discriminative models.
Additionally, \cite{van2021class} employs variational autoencoders as generative classifiers to enhance the model's performance in continuous learning.

Both the two branches of DG approaches tend to utilize a discriminative linear classifier and always operate under a common assumption of unimodal distribution for each class, it lacks the capability to holistically harness valuable and various domain-specific information that exhibits multi-modality. In contrast, we introduce a novel generative classifier for DG to capture the underlying multi-modal distribution present in domain-specific data. To the best of our knowledge, this is the first work that reveals the potential of the generative paradigm for DG classifier.

\section{Methodology}

In this section, we first analyze the risk of pursuing domain invariance for DG in Sec.~\ref{sec:3.1}. Specifically, we theoretically investigate how relying on domain-invariant features leads to an increased upper bound on the target risk. We highlight the essential impact of domain-specific information in lowering the upper bound of target risk for DG.
Drawing from these insights, we then present generative classifier-driven domain generalization (GCDG) in Sec.~\ref{sec:3.2} and make discussions on the effectiveness of GCDG in Sec.~\ref{sec:3.3}. 

\noindent \textbf{Notations.} In DG, denote $X$ and $Y$ as the variables for the input and output, respectively. There are $M$ source~(seen) domains: $S = \{(X, Y)_{S_i}\sim p_i(X, Y), 1\leq i \leq M\}$ and $p_i(X, Y) \neq p_j(X, Y), 1\leq i \neq j \leq M$. As a common practice, research in DG aims to learn a robust model $h = g\circ f$, where $f: X\rightarrow Z$ is the representation function and $g:Z\rightarrow {Y}$ is the predictive function.
Note that the variational form between the conditional entropy $H(Y|X)$ and the cross-entropy loss is: $H(Y|X) = \inf_h\mathbb{E}_{p}[\ell_{CE}(h(X), Y)]$~\cite{farnia2016minimax,zhao2022fundamental}.

\subsection{Analysis on the Risk of Domain Invariance}
\label{sec:3.1}

Bui et al.~\cite{bui2021exploiting} focused on confirming the reduction of label-related information with domain-invariant features in DG. In contrast, this work  goes further to establish a connection between domain invariance and the escalation of empirical source risk, and theoretically analyze an upward shift in the upper bound of target risk of learning domain invariance.

Inspired by the empirical observation~\cite{zhao2020domain, mahajan2021domain}, we propose that \textit{reducing the distribution gap between domains may not always result in better generalization performance}. 
To gain deeper insights, we embark on a theoretical analysis to understand how reducing the distribution gap impacts the generalization capacity. 
To facilitate the analysis, we introduce the concept of \textit{information gap} of source domains as $\Delta_p:=\sum_{i \neq m^*} (I(X_i; Y) - I(X_{m^*}; Y))$, which characterizes the feature gap concerning label predictions.
$I(;)$ denotes the mutual information and $I(X_{m^*}; Y) = \min\{ I(X_1; Y), \cdots, I(X_M; Y)\} $. 
As we delve into our analysis, the information gap is a lower bound for the increased empirical source risk when learning domain-invariant features across domains. 
Consequently, the information gap represents the cost incurred when pursuing domain-invariant representation, as demonstrated by the following theorem. 
It is noteworthy that, when referring to domain-invariant representation in the context of the subsequent theorem, we specifically denote the domain-invariant joint distribution, as opposed to the domain-invariant marginal distribution or conditional distribution. 
This choice is substantiated by extensive research highlighting the superior efficacy of domain invariance on joint distributions over marginal or conditional distributions in the context of transfer learning~\cite{zhao2020domain, long2017deep, li2018deep, courty2017joint, long2024rethinking}.
\begin{theorem}
For feature extractor $f$, if features $Z_1 = f(X_1), \cdots, Z_M = f(X_M)$ across M source domains are domain-invariant, i.e., $p(Z_1, Y) = \cdots = p(Z_M, Y)$, then $\inf_{g} \sum_{i = 1}^{M}\mathbb{E}_{p_i}[\ell_{CE}(g(Z_i), Y)] - \inf_{h} \sum_{i = 1}^{M} \mathbb{E}_{p_i}[\ell_{CE}(h(X_i), Y)]
        \ge \Delta_p.$
\label{thm1}
\end{theorem}
\paragraph{Remark 1.} Theorem~\ref{thm1} posits that the optimal empirical source risk attainable with domain-invariant features is at least $\Delta_p$ greater than what can be achieved with the original input. Notably, if certain domains contain hard samples or outliers, the pursuit of domain invariance
could result in features that lack informative power regarding the output. Consequently, this could lead to an increased empirical source risk that upper bounds the target risk~\cite{ben2006analysis, ben2010theory}. \textit{Refer to the Appendix for the proof of Theorem~\ref{thm1} and Theorem~\ref{thm2}.}

Theorem~\ref{thm1} elucidates that domain-invariant features derived from the input heighten the empirical source risk. 
However, directly inputting the high-dimensional data into the classifier is impractical due to its complexity and the extraneous information for classification. The question arises: Is it feasible to leverage the valuable domain-specific information omitted by domain-invariant features?

\begin{theorem}
Let $Z = f_1(Q) \in \mathcal{Z}$ represent domain-invariant features, and $Q = f_2(X) \in \mathcal{Q}$ represent intermediate feature representations forming the Markov chain $Y \to X \to Q \to Z$. Let $\mathcal{F}_{\text{all}}$ denote the space of all measurable predictors containing the Bayes-optimal hypothesis, and let $\mathcal{H}_Q = \{g \circ f_2 \mid g \in \mathcal{G}_Q\}$ and $\mathcal{H}_Z = \{g \circ f_1 \circ f_2 \mid g \in \mathcal{G}_Z\}$ be constrained hypothesis classes such that $\mathcal{H}_Z \subseteq \mathcal{H}_Q \subseteq \mathcal{F}_{\text{all}}$. Under cross-entropy loss, the target risks $\epsilon_T(h) = \mathbb{E}_{\mathcal{D}_T}[\ell_{CE}(h(X), Y)]$ on intermediate features $Q$ and invariant features $Z$ satisfy:$$\inf_{h \in \mathcal{H}_Z} \epsilon_T(h) - \inf_{h \in \mathcal{H}_Q} \epsilon_T(h) \ge \Delta_Q,$$where $\Delta_Q = I_T(Q; Y \mid Z) \ge 0$ measures the target label information present in $Q$ that is discarded when enforcing domain invariance in $Z$.
\label{thm2}
\end{theorem}
\paragraph{Remark 2.} Theorem~\ref{thm2} indicates that pursuing domain invariance is not always the most effective strategy for DG, and relaxing the constraint of domain invariance, \emph{e.g.,} by harnessing features in the intermediate state of the domain invariance learning process, may lead to a reduction in the upper bound of the target risk.

\noindent\textbf{The Insufficiency of the Linear Classifier.} Regarding the potential of domain-specific information to enhance generalizability and its integration into DG methods, it becomes imperative to mitigate the limitations inherent in the prevailing linear classifier within DG. 
In particular, the linear classifier assigns a solitary weight vector to each class, implying an underlying presumption of data's unimodality within each class. 
However, this assumption tailored for domain invariance is often not applicable for multi-modal distributions in DG, which restricts the model from effectively accommodating diverse domain-specific information, \emph{i.e.,} distribution of each class exhibiting multi-modality across domains, as shown in Fig.~\ref{blue_print}(a). 
Besides, as a discriminative learning paradigm, the discriminative classifier primarily focuses on defining decision boundaries rather than modeling feature distributions, thereby limiting its ability to effectively leverage valuable domain-specific information.

To address the limitations of discriminative linear classifiers in DG, the generative paradigm presents a more suitable alternative, as it can effectively model various underlying feature distributions. 
Some DG methods~\cite{murkute2021domain, wang2022variational} have employed the generative paradigm. However, these methods merely utilize generative modeling as an auxiliary mechanism to impose additional constraints alongside the classification loss, such as image reconstruction or generation. Moreover, they remain focused on learning domain invariance rather than utilizing generative modeling to capture various underlying feature distributions.
In contrast, we introduce the generative paradigm into the DG classifier to effectively model the diverse domain-specific information, thereby reducing the upper bound of the target risk and enhancing generalization.
To the best of our knowledge, this is the first work that studies and overcomes the insufficiency of the prevalent linear classifier in DG. 

\subsection{Generative Classifier-driven Domain Generalization}
\label{sec:3.2}
In light of the detrimental impact of the domain invariance and the limitations of the linear classifier in handling diverse distributions in DG, we propose a novel approach, namely Generative Classifier-driven Domain Generalization~(GCDG), to replace the prevalent discriminative linear classifier with the proposed generative classifier. 
The goal is to relax the alignment constraints and accordingly enhance the classifier's expressiveness capability through a generative paradigm. 

\begin{figure*}
    \centering
    \includegraphics[width=0.95\textwidth]{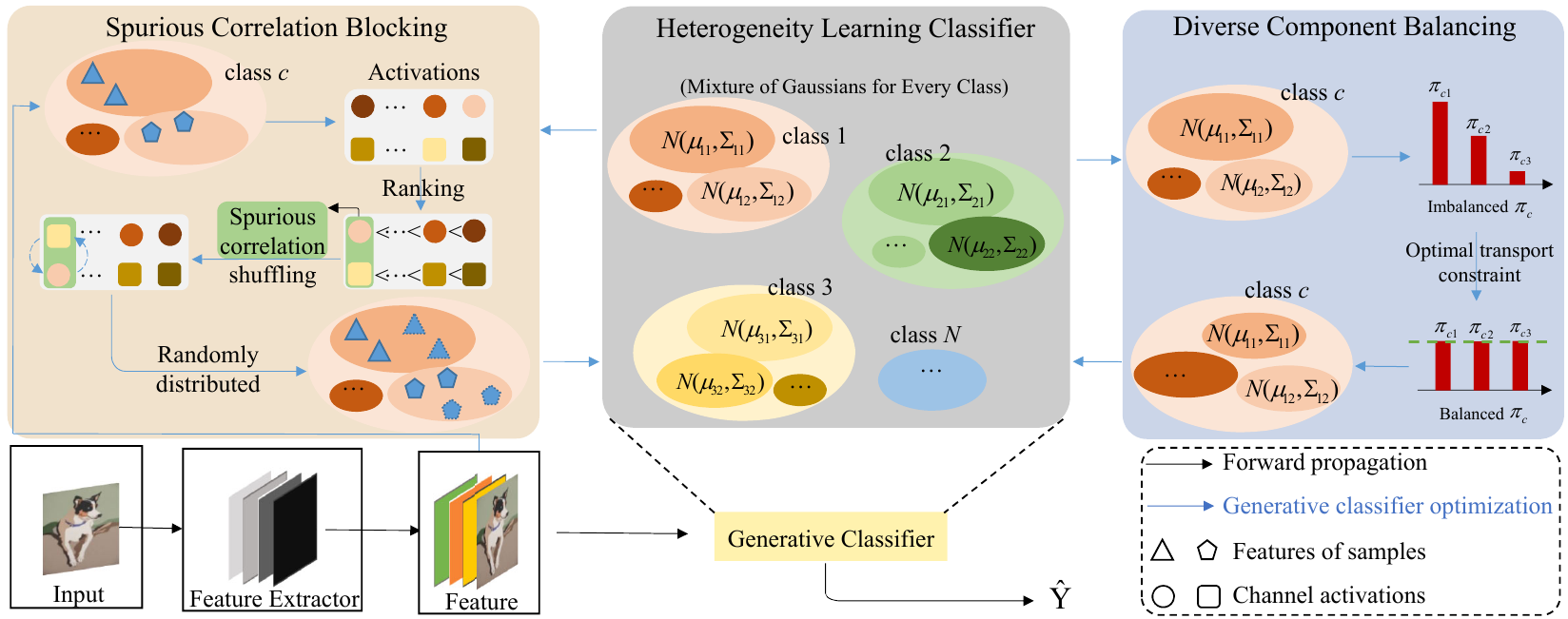}
    \caption{The framework of our proposed GCDG. The key innovation is the Heterogeneity Learning Classifier, which is a generative classifier consisting of a mixture of Gaussians for each class and adept at effectively harnessing valuable domain-specific information exhibiting multi-modality. 
    Besides, we introduce Spurious Correlation Blocking to shuffle the neural units containing spurious correlations, mitigating their adverse effect on capturing domain-specific information. 
    Furthermore, Diverse Component Balancing is designed to balance the contributions of diverse components, avoiding underestimating essential ones.
    }
    \label{framework}
    \vspace{-2mm}
\end{figure*}

Fig.~\ref{framework}
illustrates the overall architecture of GCDG, which comprises three key modules: Heterogeneity Learning Classifier (HLC), Spurious Correlation Blocking (SCB), and Diverse Component Balancing (DCB). Specifically, HLC introduces a generative classifier to capture valuable domain-specific information, thereby enhancing generalization performance. Meanwhile, SCB mitigates the adverse effects of spurious correlations on HLC’s learning process. Additionally, DCB ensures a balanced contribution of diverse components within HLC, preventing the underestimation of essential components.

\subsubsection{Heterogeneity Learning Classifier}

The key to GCDG lies in our proposed Heterogeneity Learning Classifier~(HLC), which leverages the Gaussian Mixture Model (GMM) for each class across domains, being able to model underlying multi-modal distributions and assign probabilities to different modes by using a mixture of Gaussians. 
This flexibility enables our generative classifier to excel in capturing the nuances of domain-specific information characterized by diverse distributions. 
As such, our proposed model can harness domain-specific information and thereby becomes more tolerant of intra-class variations.

Specifically, the proposed HLC adopts a mixture of $K$ Gaussians to model the diverse feature distributions of class $c$ across domains in the $D$-dimensional space:
\begin{equation}
\begin{split}
    p\left(f(x) \mid c, \phi_c\right)=&\sum_{i=1}^K p\left(i \mid c, \pi_c\right) p\left(f(x) \mid c, i, \mu_c, \Sigma_c\right) \\
    =&\sum_{i=1}^K \pi_{c i} \mathcal{N}\left(f(x)|\mu_{c i}, \Sigma_{c i}\right),
\end{split}
\label{gaussian}
\end{equation}
where $\pi_{ci} = p\left(i \mid c, \pi_c\right)$ is the mixing coefficient of component $i$, satisfying $\sum_i \pi_{ci} = 1$. $\mu_{c i} \in \mathbb{R}^D$ and $\Sigma_{c i}\in \mathbb{R}^{D\times D}$ are the mean and covariance for component $i$, respectively. We denote the parameters $\{\pi_c, \mu_c, \Sigma_c\}$ as $\phi_c$ for class $c$. 

To optimize the GMMs in our proposed HLC, the direct idea is to utilize the Expectation-Maximization (EM) algorithm, involving the E-step to evaluate the component responsibilities and the M-step to update the parameters. 
However, applying the EM algorithm to large datasets in DG is impractical due to the requirement of processing all data simultaneously for parameter updates. 

To facilitate the EM algorithm in large datasets, the semantic segmentation approach, GMMSeg~\cite{liang2022gmmseg}, employs the SK-EM algorithm~\cite{cuturi2013sinkhorn} to optimize GMMs. Additionally, to accommodate the need for parameter updates, GMMSeg introduces a feature bank to expand the sample pool for the EM algorithm.
However, the effectiveness of GMMSeg heavily relies on the sufficiency of features within the feature bank. This dependency makes GMMSeg vulnerable in DG scenarios where certain classes suffer from data scarcity, such as in OfficeHome\cite{venkateswara2017deep} and DomainNet\cite{peng2019moment}.
Table~\ref{GMMseg} reports the generalization performance of GMMseg on these datasets. As observed, ERM without any DG techniques even outperforms GMMseg. Besides, GMMSeg struggles to achieve effective convergence on DomainNet, where both the dataset scale and the number of classes are significantly larger. These findings demonstrate the limitations of GMMseg's application in DG.
To overcome these issues, we adopt the gradient descent method to optimize the GMM parameters in HLC effectively, eliminating the reliance on large-scale feature banks and enhancing performance.

\begin{table}[]
        \caption{Comparison of the generalizability on datasets where the number of samples in one class is small.} 
    \centering
    \begin{tabular}{c|c|c}
    \toprule
        Methods& OfficeHome ($\uparrow$) & DomainNet ($\uparrow$) \\
         \midrule
         ERM & 60.51 & 43.68  \\
         GMMSeg & 60.16 & 13.16 \\
         GCDG (ours) & 64.49 & 46.60 \\
         \bottomrule
    \end{tabular}
    \label{GMMseg}
    \vspace{-5mm}
\end{table}

\subsubsection{Spurious Correlation Blocking}

\begin{figure}[t]
    \centering
    \includegraphics[width=0.8\linewidth]{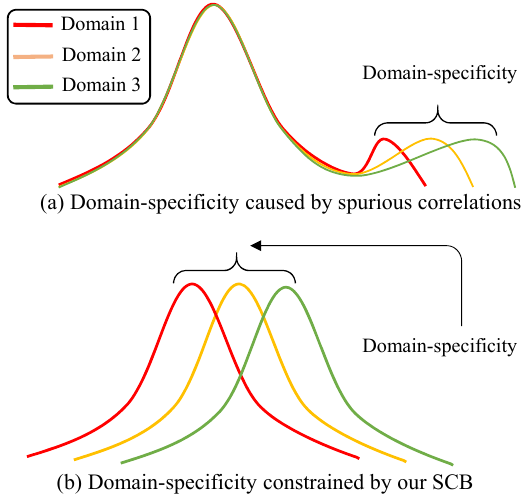}
    \caption{(a) Spurious correlations across diverse scenarios may appear as domain-specific information and be mistakenly captured by our proposed HLC, damaging the generalization performance. (b) We introduce Spurious Correlation Blocking (SCB) to perturb these spurious correlations, alleviating their detrimental effect on HLC.}
    \label{SCB-motivation}
    \vspace{-3mm}
\end{figure}

As Theorem~\ref{thm1} suggests, appropriately incorporating valuable domain-specific information can enhance generalization performance in unseen environments. However, indiscriminately leveraging domain-specific information may have adverse effects, as spurious correlations can also manifest as domain-specific features that should be suppressed to learn robust representations. As illustrated in Fig.~\ref{SCB-motivation}~(a), spurious correlations can vary across domains, and it makes no sense to capture these domain specificities as they contribute little to the accurate classification tasks. In contrast, our proposed HLC is designed to learn genuine correlations embedded within domain-specific information, as depicted in Fig.~\ref{SCB-motivation}~(b). Therefore, it is crucial to effectively isolate and suppress spurious correlations while capturing beneficial domain-specific information.

To effectively mitigate the adverse impact of spurious correlations on valuable domain-specific information, we propose Spurious Correlation Blocking (SCB). SCB is designed to identify neural units that convey spurious correlations and subsequently perturb them. Specifically, for samples within a given class, we detect spurious correlations based on feature-level activation values, which can be formulated as:
\begin{equation}
    a = -diag(\log \Sigma_{ci}) -\frac{1}{2}(z-\mu_{ci})^T(z-\mu_{ci})^T\Sigma_{ci},
\end{equation}
where $c$ and $i$ represent the class index and component index to which the samples belong, respectively. Notably, we assume that random variables in the components are independent.

The feature-level activation value quantifies the significance of different feature dimensions in contributing to classification. Typically, neural units with lower activation values are less relevant to the classification task and are thus more likely to encode spurious correlations, which can hinder the effectiveness of our proposed heterogeneity learning. In contrast, units with higher activation values capture meaningful information that could enhance generalization performance.
To mitigate the adverse effects of spurious correlations, our proposed SCB shuffles the units with lower activation values within the same class, ensuring that spurious correlations are randomly distributed and thereby blocking their detrimental influence. Mathematically, for samples $m$ and $n$, the shuffling process in SCB can be formulated as:
\begin{equation}
\begin{aligned}
    z_m^{SCB} = z_m \otimes (1-M_m) + z_n \otimes M_n, \\
    z_n^{SCB} = z_n \otimes (1-M_n) + z_m \otimes M_m, 
    \label{SCB}
\end{aligned}
\end{equation}
where $\otimes$ denotes the element-wise multiplication, and $M$ represents the selecting mask used to identify neural units conveying spurious correlations. For sample $m$, the selecting mask $M_m$ could be formulated as:
\begin{equation}
    M_{m, i} = \left\{
    \begin{aligned}
    0&, & z_{m, i} \leq Q_q(z_{m})\\
    1 &, & z_{m, i} > Q_q(z_{m})
    \end{aligned}
    \right.,
\end{equation}
where $Q_q$ denotes the $q$-th percentile for $z_{m}$.

\subsubsection{Diverse Component Balancing}

The proposed HLC and SCB enable the model to effectively leverage critical domain-specific information that enhances generalization performance. However, within a GMM for a given class, the contributions of different components may vary due to the imbalance of data across domains. 
Consequently, essential domain-specific information associated with a small subset of data may be underestimated or even ignored, thereby limiting its potential to improve generalizability.

To mitigate the adverse effects of data imbalance on unbalanced component contributions in GMM and to prevent degenerate solutions in Eq.~\eqref{gaussian}, where the model overlooks essential domain-specific information, causing the mixing coefficients of certain components to become negligible, we impose a constraint requiring the mixing coefficients to follow a uniform distribution:
\begin{equation}
    \pi_{ci} = \frac{1}{K} = \frac{\sum_n \gamma_{cni}}{N_c},
\end{equation}
where $\gamma_{cni}$ is the posterior of component $i$ for sample $n$ in class $c$, and $N_c$ denotes the number of samples in class $c$. Combined with the characteristic of the posterior $\gamma_{cni}$, the following constraints hold:
\begin{equation}
        \sum_n \gamma_{cni} = \frac{N_c}{K}, \quad \sum_i \gamma_{cni} = 1.
        \label{condition}
\end{equation}
 
The conditions in Eq.~\eqref{condition} are combinational in the posterior $\gamma_{cni}$ and thereby challenge to optimize.
To this end, we adopt entropic optimal transport~\cite{mena2020sinkhorn, liang2022gmmseg} to facilitate the computation of the feature posterior $\Gamma$ of $N_c$ samples:
\begin{equation}
\begin{aligned}
\min _{\Gamma_c} \quad &\Gamma_c \otimes O_c+\lambda H\left(\Gamma_c\right), \\
\mathrm{s.t.} \ \ \Gamma_c \in \mathbb{R}_{+}^{N_c \times K},
 &\Gamma_c \mathbf{1}^K=\mathbf{1}^{N_c},
 \left(\Gamma_c\right)^{\top} \mathbf{1}^{N_c}=\frac{N_c}{K} \mathbf{1}^K,\\
 \label{sk-E}
\end{aligned}
\end{equation}
where $\otimes$ means the element-wise multiplication, and $\Gamma_c$ represents the posterior matrix with $\Gamma_c(n, i) = \gamma_{cni}$. Additionally, $O_c$ is the cost matrix, and $O_c(n, i) = -\log p(f_{cn}(x)|c, i)$. $\lambda$ denotes the Lagrange multiplier, $H(\cdot)$ is the entropy function,
and $\mathbf{1}^K$ is a $K$-dimensional all-one vector. 
As indicated by the Sinkhorn-Knopp algorithm in ~\cite{cuturi2013sinkhorn, asano2019self}, the solution to Eq.~\eqref{sk-E} can be formulated as:
\begin{equation}
    \Gamma_c^{\ast} = \text{diag}(a) \exp(-\lambda O_c)\text{diag}(b),
    \label{solution}
\end{equation}
where $a$ and $b$ are two scaling vectors ensuring that the transport matrix $\Gamma_c$ presents a probability matrix. The optimization of $a$ and $b$ is performed through the following iterations:
\begin{equation}
    a_i = (\exp(-\lambda O_c)b_{i-1})^{-1}, \quad b_i = (a_{i-1}^T\exp(-\lambda O_c))^{-1},
    \label{a_b}
\end{equation}
where $i$ denotes the iteration number and is set to 3 in all experiments unless specified.

\begin{algorithm}[tb]
\caption{Training algorithm of our GCDG}
\label{algorithm}
\textbf{Input}: $M$ source domains: $\{S_i\}_{i = 1}^{M}$, feature extractor $f$, generative classifier $g$\\
\textbf{Parameter}: Number of components: $K$, quantile parameter: $q$, Lagrange multiplier: $\lambda$\\
\textbf{Output}: the generative hypothesis model $h = g \circ f$
\begin{algorithmic}[1] 
\While{training is not converged}
\State Sample data from $S$
\State Obtain the features by forwarding the samples through the feature extractor $f$
\State Shuffle the features by Eq.~\eqref{SCB}
\State Update the scaling vectors $a$ and $b$ by Eq.~\eqref{a_b}
\State Seek the feature posterior $\Gamma$ by Eq.~\eqref{solution}
\State Predict the result by the generative classifier $g$
\State Calculate the prediction loss and optimize the feature extractor $f$ and the classifier $g$
\EndWhile
\end{algorithmic}
\end{algorithm}

The pseudo-code of the optimization for our proposed GCDG is demonstrated in Algorithm~\ref{algorithm}.

\subsection{Discussion on the Effectiveness of GCDG}

\label{sec:3.3}
In this section, we analyze the effectiveness of GCDG and how it promotes generalization performance, drawing valuable insights from the theoretical results in DG.

\noindent\textbf{Lowering the Bound for the Target Risk.} Combined with the bound on target risk~\cite{ben2006analysis} in domain adaptation, Theorem~\ref{thm2} highlights that we can effectively decrease the upper bound of target risk by incorporating extra information instead of solely relying on domain-invariant features. Building upon this insight, we leverage the expressive power of the generative classifier to model diverse data distributions across domains by combining multiple Gaussians. This approach enables us to capture a wide range of data patterns and mitigate the loss of valuable domain-specific information for output prediction, consequently reducing the source risk as well as the upper bound of the target risk.

To verify our claim that GCDG can reduce source risk, we report its in-domain performance across five benchmarks in Table~\ref{in-domain}. As observed, GCDG surpasses flat-minima-seeking methods that could minimize source risk and enhance in-domain performance in DG, demonstrating its effectiveness in both aspects. Consequently, by relaxing alignment constraints and accommodating essential domain-specific information, as indicated by Theorem~\ref{thm2}, GCDG facilitates a reduction in the upper bound of target risk, thereby improving generalization.

\begin{table}[]
    \centering
    \caption{Comparisons of in-domain generalization on five DG benchmarks with flatness-aware optimization methods.}
    \setlength{\tabcolsep}{3pt}
    \resizebox{0.49\textwidth}{!}{
    \begin{tabular}{c|ccccc|c}
    \toprule
     Model & PACS ($\uparrow$) & VLCS ($\uparrow$) & OH ($\uparrow$) & TI ($\uparrow$) & DN ($\uparrow$) & Avg. ($\uparrow$) \\
     \midrule
     SAM \cite{foret2021sharpnessaware}                  & 96.64 & 85.01 & 79.30 & 91.25 & 64.44& 83.33\\
     SWAD \cite{cha2021swad}                  & 96.20 & 84.44 & 78.53 & 90.90 & 64.44 & 82.90\\
     PCL \cite{yao2022pcl}                    & 96.17 & 84.16 & 79.60 & 87.89 & 64.25 & 82.41\\
     \midrule
     GCDG (ours)                           & \bfseries96.98 & \bfseries85.52 &\bfseries79.62 & \bfseries92.75 & \bfseries64.91 & \bfseries83.96\\
     \bottomrule
    \end{tabular}
    }
    \label{in-domain}
    \vspace{-5mm}
\end{table}

\noindent\textbf{Promoting Flat Minima.} The pursuit of flat minima in the loss landscape has been acknowledged for its potential to enhance generalization performance, as it renders the model less susceptible to small input data perturbations~\cite{izmailov2018averaging, he2019asymmetric, foret2021sharpnessaware}. The notion of flat minima has garnered considerable attention in transfer learning to promote generalization performance~\cite{kim2021selfreg,cha2021swad, wang2023sharpness}. Cha et al. \cite{cha2021swad} theoretically implied that seeking flat minima
can reduce the domain generalization gap on target domains.

We emphasize that our GCDG aligns with the pursuit of flat minima. Numerous works seek to increase posterior entropy~\cite{zhang2019your, zhang2018deep}, allowing the model to converge to flatter minima by accommodating more information encoded in soft labels during training. 
In contrast, approaches that force the model to fit samples experiencing distribution shifts to identical one-hot labels can lead to convergence to less flat minima, making the model more sensitive to small perturbations. 
In this context, we demonstrate that GCDG induces higher entropy in the feature space by capturing diverse features, as evidenced in Table~\ref{entropy}.
The capability of GCDG to handle multimodal distribution, rather than compelling the model to solely capture domain-invariant features, relaxes the training procedure akin to the principles of entropy regularization methods~\cite{zhang2018deep}.

To visually illustrate the superiority of our proposed GCDG in promoting flat minima, Fig.~\ref{visual_swad} in Section~\ref{section_5} presents a direct visualization of the loss landscapes of different methods, including SWAD~\cite{cha2021swad}, which explicitly seeks flat minima through dense stochastic weight averaging. 
Notably, GCDG exhibits a stronger capability in achieving flat minima, highlighting its effectiveness in enhancing generalization. This result underscores the advantage of GCDG in promoting flat minima by accommodating diverse feature distributions, thereby improving generalizability.

\begin{table}[]
        \caption{Comparison of the average entropy values of features on source domains when the model is converged.} 
    \centering
    \begin{tabular}{c|cc|cc}
    \toprule
        \multirow{2}{*}{Methods}& \multicolumn{2}{c|}{Office-Home (Clipart)}& \multicolumn{2}{c}{PACS (Cartoon)}\\
        & Entropy & Accuracy & Entropy & Accuracy\\
         \midrule
         ERM & 7.04 & 48.00 & 7.96 & 74.79\\
         GCDG & \bfseries7.62 & \bfseries51.27& \bfseries8.66 & \bfseries78.58 \\
         \bottomrule
    \end{tabular}
    \label{entropy}
    \vspace{-3mm}
\end{table}

 \section{Experiments}

\subsection{Experiment Details}
\noindent \textbf{Dataset.} Following previous DG protocols~\cite{gulrajani2020search}, we compare our GCDG with state-of-the-art methods on five benchmarks: (1) PACS~\cite{li2017deeper} consists of 9991 images categorized into 7 classes from 4 styles. (2) VLCS~\cite{fang2013unbiased} contains four datasets, including 10729 images distributed in 5 categories. (3) Office-Home~\cite{venkateswara2017deep} comprises 15588 images in 65 categories of 4 datasets. (4) Terra-Incognita~\cite{beery2018recognition} consists of 24330 photographs of 10 kinds of wide animals taken at 4 locations. (5) DomainNet~\cite{peng2019moment}, presenting a greater challenge for DG, includes 586575 images in 345 classes from 6 domains.

\noindent\textbf{Evaluation Metrics.} For a fair comparison, we adopt the training-domain validation following previous DG protocols~\cite{gulrajani2020search, li2023sparse, wang2023sharpness}, choosing one domain as the target domain and training on the remaining domains. We split samples from each source domain to an 8:2 ratio for training and validation, respectively. For the performance, we take turns selecting a domain as the target domain, then report the accuracy on each target domain and their average.

\noindent\textbf{Network Architecture.} We adopt ResNet-18, ResNet-50, and DeiT-S as feature extractors. For methods whose original papers did not report results on ResNet‑18 or provide per‑domain details, we present our reproduced results. Before forwarding features into the generative classifier, we apply a fully connected layer to reduce the feature dimensionality to $D$, facilitating the generative classifier updating and inference. To further improve computational efficiency, we enforce the covariance matrix $\Sigma \in \mathbb{R}^{D\times D}$ to be diagonal for all components. It is worth noting that the generative classifier upon the feature extractor is a versatile module for not only the models but also the backbones in DG.
\begin{table}[h]
    \centering
    \caption{Hyperparameter search space.}
    \begin{tabular}{c|c}
    \toprule
         Parameter & Value  \\
         \hline
         Learning rate & [5e-5, 8e-5]\\
         Number of components $K$ & [2, 3, 5]\\
         Compression dimension $D$  & [64, 1024] \\
         Quantile $q$ & [10, 20, 30]\\
         \bottomrule
    \end{tabular}
    \label{hyper}
    \vspace{-3mm}
\end{table}

\begin{table}
    \centering
    \caption{Generalization results of state-of-the-art methods and our GCDG on PACS.}
    \setlength{\tabcolsep}{4pt}
    \resizebox{0.49\textwidth}{!}{%
    \begin{tabular}{r|cccc|c}
    \toprule
    \multirow{2}{*}{Method} & \multicolumn{4}{c|}{Target domain} & \multirow{2}{*}{Avg.($\uparrow$)}\\
    \cmidrule(r){2-5}
     & Art & Cartoon & Photo & Sketch \\
    \midrule
    \multicolumn{6}{c}{ResNet-18} \\
    \midrule
    
    RSC \cite{huang2020self} (ECCV'2020)      & 79.88         & 76.87        & 94.56         & 77.11        & 82.10 \\
    MTL \cite{blanchard2021domain} (JMLR'2021)  & $79.99$         & $72.18$        & $95.28$         & $74.94$        & $80.60$  \\
    SagNet \cite{nam2021reducing} (CVPR'2021)  & 81.15           & $75.05$        & $94.61$         & 75.38          & $81.55$ \\
    ARM \cite{zhang2021adaptive} (NeurIPS'2021)     & $80.42$         & $75.96$        & $95.21$         & $72.33$        & $80.98$ \\
    SAM \cite{foret2021sharpnessaware} (ICLR'2021)    & $80.67$         & $75.53$        & 93.86         & $79.33$        & $82.35$ \\
    SWAD \cite{cha2021swad} (NeurIPS'2021)    & $83.28$         & $74.63$        & 96.56         & $77.96$        & $83.11$ \\
    PCL \cite{yao2022pcl} (CVPR'2022)     & 83.53           & 73.61          & $96.18$         & $77.20$        & $82.63$\\
    AdaNPC \cite{zhang2023adanpc} (ICML'2023) & 82.70           & 76.80          & 92.80         & 77.70        & 82.50\\
   
    SAGM \cite{wang2023sharpness} (CVPR'2023) & 81.76           & 74.68          & 95.51         & 73.41        & 81.34\\
    iDAG \cite{huang2023idag} (ICCV'2023) & 82.18           & 78.20          & 97.08         & 75.38       & 83.21\\
    GMDG \cite{tan2024rethinking} (CVPR'2024)& \bfseries83.77  &75.64 & \bfseries97.38 & 67.91 & 81.71\\
    SDCL \cite{li2026causal} (TPAMI'2026)& 78.77  &75.69 &93.71 & 75.86 & 81.01\\
    CFSM \cite{zhu2026cfsm} (TPAMI'2026)& 83.22  &77.67 &95.58 & 75.76 & 83.06\\
\rowcolor{lightgray}    GCDG (ours) & 83.06           & \bfseries78.50 & 92.63         & \bfseries79.56 & \bfseries83.44 \\ 
\midrule
    \multicolumn{6}{c}{DeiT-S} \\
    \midrule   
    SDViT \cite{sultana2022self} (ACCV'2022) & 87.60          & 82.40          & 98.00         & 77.20       & 86.30\\
    GMoE \cite{li2023sparse} (ICLR'2023) & \bfseries89.40           & 83.90         & \bfseries99.10         & 74.50       & 86.70\\
    DTCS \cite{long2025diverse} (TIP'2025)& 90.58  &82.94 & 98.86 & 73.22 & 86.40\\
   \rowcolor{lightgray}    GCDG (ours) & 88.60          & \bfseries85.60 & 98.60         & \bfseries79.30 & \bfseries88.00 \\ 
   \bottomrule
    \end{tabular}
    }
    \label{PACS}

\end{table}
 
\begin{table}[t]
    \centering
    \caption{Performance comparison with state-of-the-art methods on Terra-Incognita.}
    \setlength{\tabcolsep}{4pt}
    \resizebox{0.47\textwidth}{!}{%
    \begin{tabular}{r|cccc|c}
    \toprule
    \multirow{2}{*}{Method} & \multicolumn{4}{c|}{Target domain} & \multirow{2}{*}{Avg.($\uparrow$)}\\
    \cmidrule(r){2-5}
     & L100 & L38 & L43 & L46 \\
    \midrule
    \multicolumn{6}{c}{ResNet-18} \\
    \midrule
   
    RSC \cite{huang2020self} (ECCV'2020)    & $47.32$         & $37.66$        & $51.67$         & 35.95          & $43.15$\\
   
    MTL \cite{blanchard2021domain} (JMLR'2021)    & $38.94$         & $35.18$        & $52.80$         & $35.29$        & $40.55$ \\
    SagNet \cite{nam2021reducing} (CVPR'2021) & 47.25           & 29.67          & 52.87           & 25.22          & $38.75$\\
    ARM \cite{zhang2021adaptive} (NeurIPS'2021)    & $44.98$         & $33.73$        & $43.39$         & $27.77$        & $37.47$\\
    SAM \cite{foret2021sharpnessaware} (ICLR'2021)   & \bfseries55.66         & 27.92        & 51.51  & 31.93       & 41.76\\
    SWAD \cite{cha2021swad} (NeurIPS'2021)   & $49.80$         & $33.16$        & \bfseries55.57  & $33.19$        & $42.93$\\
   
    PCL \cite{yao2022pcl} (CVPR'2022)   & $52.62$         & $39.98$        & $48.49$         & $31.74$        & $43.21$\\
    AdaNPC \cite{zhang2023adanpc} (ICML'2023) & 50.60           & 38.60         & 42.20         & 34.00        & 41.35\\
    
    SAGM \cite{wang2023sharpness} (CVPR'2023) & 50.20           & 27.54         & 53.21         & 31.70        & 40.66\\
    iDAG \cite{huang2023idag} (ICCV'2023) & 53.78           & 34.82         & 50.28         & 28.85        & 41.93\\
  GMDG \cite{tan2024rethinking} (CVPR'2024)& 50.70  & 34.78 & 51.26 & 36.63 & 43.34\\
  SDCL \cite{li2026causal} (TPAMI'2026)& 50.70  &36.86 & 47.29 & 31.57 & 41.61\\
    CFSM \cite{zhu2026cfsm} (TPAMI'2026)& 47.59  &41.58 &53.56 & 35.10 & 44.46\\
   \rowcolor{lightgray} GCDG (ours)& 49.23          & \bfseries41.96          & 51.71         & \bfseries36.56          & \bfseries44.86 \\
    \midrule
    \multicolumn{6}{c}{DeiT-S} \\
    \midrule  
    SDViT \cite{sultana2022self} (ACCV'2022)& 55.90          & 31.70          & \bfseries52.20         & 37.40       & 44.30\\
    GMoE \cite{li2023sparse} (ICLR'2023) & 59.20          & 34.00         & 50.70         & 38.50       & 45.60\\
    DTCS \cite{long2025diverse} (TIP'2025)& 50.31  &26.02 & 54.01 & 38.93 & 42.32\\
     \rowcolor{lightgray} GCDG (ours) & \bfseries59.20          & \bfseries35.79 & 50.45         & \bfseries39.10 & \bfseries46.12 \\ 
    \bottomrule
    \end{tabular}
    }
    \label{terra}
    \vspace{-5mm}
\end{table}

\begin{table}[t]
    \centering
    \caption{Performance comparison with state-of-the-art approaches on Office-Home.}
    \setlength{\tabcolsep}{4pt}
    \resizebox{0.487\textwidth}{!}{%
    \begin{tabular}{r|cccc|c}
    \toprule
    \multirow{2}{*}{Method} & \multicolumn{4}{c|}{Target domain} & \multirow{2}{*}{Avg.($\uparrow$)}\\
    \cmidrule(r){2-5}
     & Art & Clipart & Product & Real \\
    \midrule
    \multicolumn{6}{c}{ResNet-18} \\
    \midrule
    
    RSC \cite{huang2020self} (ECCV'2020)      & $49.38$         & $45.91$        & $66.84$         & $67.41$         & $57.38$ \\
    
    MTL \cite{blanchard2021domain} (JMLR'2021)      & $52.58$         & $46.99$        & $70.83$         & $72.46$         & $60.72$  \\
    SagNet \cite{nam2021reducing} (CVPR'2021)   & 56.28           & 51.32          & $70.64$         & 73.38           & $62.90$\\
    ARM \cite{zhang2021adaptive} (NeurIPS'2021)      & $52.68$         & $45.82$        & $68.64$         & $71.40$         & $59.63$ \\
    SAM \cite{foret2021sharpnessaware} (ICLR'2021)     & 53.09        & 49.28        & 69.37        & 72.40        & 61.04 \\
    SWAD \cite{cha2021swad} (NeurIPS'2021)     & $54.33$         & $49.80$        & $70.92$         & $71.97$         & $61.75$ \\
    
    PCL \cite{yao2022pcl} (CVPR'2022)     & $56.69$         & 52.49 & 72.24         & 74.50        & $63.98$\\
    AdaNPC \cite{zhang2023adanpc} (ICML'2023) & 54.40           & 48.40         & 67.40         & 68.60        & 59.70\\
   
    SAGM \cite{wang2023sharpness} (CVPR'2023) & 54.22           & 49.49         & 70.61         & 73.09        & 61.85\\
    iDAG \cite{huang2023idag} (ICCV'2023) & 54.53           & 48.77         & 71.71         & 74.84        & 62.46\\
    GMDG \cite{tan2024rethinking} (CVPR'2024)& 56.23  & 50.20 & \bfseries73.34 & \bfseries75.30 & 63.77\\
    SDCL \cite{li2026causal} (TPAMI'2026)& 52.06  &53.06 & 70.19 &69.99 & 61.33\\
    CFSM \cite{zhu2026cfsm} (TPAMI'2026)& 55.66  &49.97 &73.17 & 75.01 & 63.46\\
   \rowcolor{lightgray} GCDG(ours) & \bfseries58.84  & \bfseries52.51          & 72.22  & 74.39           & \bfseries64.49\\
    \midrule
    \multicolumn{6}{c}{DeiT-S} \\
    \midrule
    SDViT \cite{sultana2022self} (ACCV'2022) & 68.30          & 56.30          & 79.50         & 81.80       & 71.50\\
    GMoE \cite{li2023sparse} (ICLR'2023) & 69.30           & \bfseries58.00         & 79.80         & 82.60       & 72.40\\
    DTCS \cite{long2025diverse} (TIP'2025)& 65.10  &55.76 & 78.08 & 80.61 & 69.89\\
    \rowcolor{lightgray} GCDG(ours) & \bfseries69.60          & 57.40 & \bfseries80.30         & \bfseries82.80 & \bfseries72.50 \\ 
    \bottomrule
    \end{tabular}   
    }
    \label{office_home}
\end{table}

\begin{table}[t]
    \centering
    \caption{Generalization results of state-of-the-art methods and our GCDG on VLCS.}
    \setlength{\tabcolsep}{3pt}
    \resizebox{0.503\textwidth}{!}{
    \begin{tabular}{r|cccc|c}
    \toprule
    \multirow{2}{*}{Method} & \multicolumn{4}{c|}{Target domain} & \multirow{2}{*}{Avg.($\uparrow$)}\\
    \cmidrule(r){2-5}
     & Caltech & LabelMe & SUN & PASCAL \\
    \midrule
    \multicolumn{6}{c}{ResNet-18} \\
    \midrule
    
    RSC \cite{huang2020self} (ECCV'2020)    & $93.29$         & $64.47$        & $71.52$         & $73.31$        & $75.65$\\
   
    MTL \cite{blanchard2021domain} (JMLR'2021)    & $96.38$         & $62.54$        & $70.91$         & $71.68$        & $75.38$ \\
    SagNet \cite{nam2021reducing} (CVPR'2021)  & 97.09           & $62.07$        & $70.37$         & 75.42          & 76.24\\
    ARM \cite{zhang2021adaptive} (NeurIPS'2021)    & $96.29$         & $61.55$        & $72.32$         & $76.27$        & $76.61$\\
    SAM \cite{foret2021sharpnessaware} (ICLR'2021)        & \bfseries98.15         & 60.52         & 71.25          & 75.90 & 76.45\\
    SWAD \cite{cha2021swad} (NeurIPS'2021)        & 97.70           & 61.27          & 70.72           & \bfseries76.71 & 76.60\\
   
    PCL \cite{yao2022pcl} (CVPR'2022)         & 97.09           & 62.07          & 71.06           & 75.05          & 76.32\\
     AdaNPC \cite{zhang2023adanpc} (ICML'2023) & 98.00           & 60.20        & 69.10         & 76.60        & 75.98\\
   
     SAGM \cite{wang2023sharpness} (CVPR'2023) & 96.03           & 60.99        & 70.64         & 75.68        & 75.83\\
     iDAG \cite{huang2023idag} (ICCV'2023) & 94.44           & 59.88        & 70.18         & 72.86        & 74.34\\
      GMDG \cite{tan2024rethinking} (CVPR'2024)& 96.56  & \bfseries63.53 & 69.35 & 73.83 & 75.81\\
       SDCL \cite{li2026causal} (TPAMI'2026)& 97.17  &60.14 & 68.96 & 72.38 & 74.67\\
    CFSM \cite{zhu2026cfsm} (TPAMI'2026)& 96.29  &61.32 &72.77 & 74.23 & 76.15\\
   \rowcolor{lightgray} GCDG (ours)     & 96.68           & 63.40          & \bfseries72.61           & 74.76          & \bfseries76.86 \\   
    \midrule
    \multicolumn{6}{c}{DeiT-S} \\
    \midrule
    SDViT \cite{sultana2022self} (ACCV'2022) & 96.80          & 64.20         & 76.20         & 78.50       & 78.90\\
    GMoE \cite{li2023sparse} (ICLR'2023) & 96.90           & 63.20         & 72.30         & \bfseries79.50       & 78.00\\
    DTCS \cite{long2025diverse} (TIP'2025)& 96.54  &64.80 & 75.75 & 77.61 & 78.67\\
   \rowcolor{lightgray} GCDG (ours) & \bfseries97.70          & \bfseries64.40 &\bfseries76.70         & 78.40 & \bfseries79.30\\ 
    \bottomrule
    \end{tabular}
    }
    \vspace{-5mm}
    \label{VLCS}
\end{table}

\noindent\textbf{Training.} We train the model for 5k iterations. 
For training, we initialize the backbone with ResNet-18 or DeiT-S pre-trained on ImageNet~\cite{deng2009imagenet}, and optimize the feature extractor and the generative classifier using Adam optimizer. For DomainNet which is a complex dataset, undergoing 5k iterations proves inadequate for achieving effective model convergence. Therefore, following the strategy in recent state-of-the-art research~\cite{cha2021swad, li2023sparse}, we utilize pre-trained ResNet-50 or DeiT-S architectures, and optimize the proposed model for 15k iterations with Adam optimizer. The learning rate is decayed by 0.1 at 60\% and 80\% of the total iterations. The batch size for each source domain is fixed at 32. 
We provide the remaining hyperparameters in Table~\ref{hyper}.

\noindent\textbf{Inference.} 
To make predictions, we simply select the component with the highest responsibility for each class.

\begin{table*}[t]
    \centering
    \caption{Comparison to the state-of-the-art FAS methods. The bold numbers indicate the best performance.} \label{fas}
\resizebox{0.98\textwidth}{!}{%
\begin{tabular}{r|cc|cc|cc|cc}
\hline \multirow{2}{*}{ \textbf{Methods} } & \multicolumn{2}{c|}{ \textbf{I\&C\&M to O} } & \multicolumn{2}{c|}{ \textbf{O\&C\&I to M} } & \multicolumn{2}{c|}{\textbf{O\&M\&I to C} } & \multicolumn{2}{c}{ \textbf{Avg.} } \\
& HTER(\%) & AUC(\%) & HTER(\%) & AUC(\%) & HTER(\%) & AUC(\%) & HTER(\%) & AUC(\%) \\
\midrule

MADDG \cite{shao2019multi} (CVPR'2019)& 27.98 & 80.02 &  17.69 & 88.06 & 24.50 & 84.51 &23.39 & 84.20 \\
D$^2$AM \cite{chen2021generalizable} (AAAI'2021)& 15.27 & 90.87 &  12.70 & 95.66 & 20.98 & 85.58 &16.32 &90.70  \\
SSDG \cite{jia2020single} (CVPR'2020)& 25.17 & 81.83 &  16.67 & 90.47 & 23.11 & 85.45 & 21.65 & 85.92 \\
RFM \cite{shao2020regularized} (AAAI'2020)& 16.45 & 91.16 &  13.89 & 93.98 & 20.27 & 88.16 & 16.87 & 91.1 \\
DRDG \cite{liu2021dual} (IJCAI'2021)& 15.63 & 91.75 &  12.43 & 95.81 & 19.05 & 88.79 & 15.70 & 92.12 \\
ANRL \cite{liu2021adaptive} (ACM MM'2021)& 15.67 & 91.90 &  10.83 & 96.75 & 17.85 & 89.26 & 14.78 & 92.64 \\
FGHV \cite{liu2022feature} (AAAI'2022)& 13.58 & 93.55 &  9.17 & 96.92 & 12.47 & 93.47 & 11.74 & 94.65 \\
SSAN \cite{wang2022domain} (CVPR'2022)& 19.51 & 88.17 &  10.42 & 94.76 & 16.47 & 90.81 & 15.47 & 91.25 \\
AMEL \cite{zhou2022adaptive} (ACM MM'2022)& 11.31 & 93.96 &  10.23 & 96.62 & 11.88 & 94.39 & 11.14 & 94.99\\
EBDG \cite{du2022energy} (ACM MM'2022)& 15.66 & 92.02 &  9.56 & \bfseries97.17 & 18.34& 90.01 & 14.52 & 93.07\\
IADG \cite{zhou2023instance} (CVPR'2023)& 11.45 & 94.50 &  8.45 & 96.99 & 12.74 & 94.00 & 10.88 & 95.16\\
EBFAS-GA \cite{zhang2024domain} (TMM'2024)& 15.56 & 92.52 &  9.69 & 96.98 & 19.34 & 89.32 & 14.86 & 92.94\\
FA\&CS \cite{cai2025towards} (IJCV'2025)& 12.10 & 94.32 &  8.04 & 96.21 & 19.50 & 87.46 & 13.21 & 92.66\\
\midrule
GCDG (ours) & \bfseries9.13 & \bfseries95.56 &  \bfseries7.50 & 96.79 & \bfseries10.92 & \bfseries94.93 & \bfseries9.18 & \bfseries95.76 \\
\bottomrule
\end{tabular}
}
\vspace{-1mm}
\end{table*}

\begin{table}
    \caption{Performance comparison with state-of-the-art methods on DomainNet. $^{\dag}$ denotes reproduced results.}
    \label{domainnet}
    \centering
    \setlength{\tabcolsep}{2pt}
    \resizebox{0.49\textwidth}{!}{%
    \begin{tabular}{r|cccccc|c}
    \toprule
    \multirow{3}{*}{Method} & \multicolumn{6}{c|}{Target domain} & \multirow{3}{*}{Avg.($\uparrow$)}\\
    \cmidrule(r){2-7}
    & Clipart & Infograph & painting & Quickdraw & Real & Sketch \\
    \midrule
    \multicolumn{8}{c}{ResNet-50}\\
    \midrule
   
    RSC \cite{huang2020self} (ECCV'2022) &55.0         & 18.3       & 44.4        & 12.2        & 55.7& 47.8 & 38.9\\
    MTL \cite{blanchard2021domain} (JMLR'2021)& 57.9         & 18.5        & 46.0         & 12.5       & 59.5 & 49.2 & 40.6 \\
    
    SagNet \cite{nam2021reducing} (CVPR'2021)& 57.7           & 19.0        & 45.3         & 12.7         & 58.1& 48.8 & 40.3 \\
    ARM \cite{zhang2021adaptive} (NeurIPS'2021)  & 49.7       &16.3       & 40.9        & 9.4      & 53.4& 43.5 & 35.5\\
    SAM \cite{foret2021sharpnessaware} (ICLR'2021)        & 64.1        & 21.1         &49.4         &14.4& 63.0 &53.7 &44.3 \\
    SWAD \cite{cha2021swad} (NeurIPS'2021) & 66.0        & 22.4       & 53.5       & 16.1      & 65.8 & 55.5 & 46.5 \\
    SWAD$^{\dag}$ \cite{cha2021swad} (NeurIPS'2021) & 66.0        & 22.2       & \bfseries53.6       & \bfseries15.4      & 65.3 & 54.8 & 46.2 \\
    PCL \cite{yao2022pcl} (CVPR'2022) & 67.9           &24.3         &55.3        &15.7       & 66.6& 56.4 & 47.7\\
    PCL$^{\dag}$ \cite{yao2022pcl} (CVPR'2022) & 64.6           &23.2         &52.9        &15.0       & 64.0 & \bfseries55.2 & 45.8\\
     AdaNPC \cite{zhang2023adanpc} (ICML'2023) & 59.3           & 22.2        & 48.3         & 14.3        & 61.0 & 51.4 & 42.8\\
    
    SAGM \cite{wang2023sharpness} (CVPR'2023) & 64.9          & 21.1          & 51.5       & 14.8       & 64.1& 53.6 & 45.0\\
    iDAG \cite{huang2023idag} (ICCV'2023)&67.9           & 24.2          & 55.0        & 16.4       & 66.1 & 56.9 & 47.7\\
    iDAG$^{\dag}$ \cite{huang2023idag} (ICCV'2023)&63.0           & 22.7          & 53.1       & 15.1      & 64.6 & 53.6 & 45.3\\
    GMDG~\cite{tan2024rethinking} (CVPR'2024)& 63.4          & 22.4          & 51.4      & 13.4       & 64.4& 52.4 & 44.6\\
    SDCL \cite{li2026causal} (TPAMI'2026)& 62.1  &21.3 &50.4 & 15.8 & 60.7 &54.3 &44.1 \\
    CFSM$^{\dag}$ \cite{zhu2026cfsm} (TPAMI'2026)& 60.4  &22.2 &52.2 & 14.6 & 62.1 &53.4 &44.2 \\
    \rowcolor{lightgray} GCDG (ours) & \bfseries66.4           & \bfseries23.8 & 53.4        & 15.0 & \bfseries66.1 &54.9 & \bfseries46.6 \\ 
\midrule
\multicolumn{8}{c}{DeiT-S}\\
    \midrule
    SDViT~\cite{sultana2022self} (ACCV'2022)  & 63.4           & 22.9          & 53.7      & 15.0       & 67.4 & 52.6 & 45.8\\
    GMoE \cite{li2023sparse} (ICLR'2023) & 68.2          & 24.7          & \bfseries55.7       & 16.3       &\bfseries69.1& 55.4&48.3 \\
    DTCS \cite{zhu2026cfsm} (TIP'2025)& 66.6  &23.6 &54.5 & 16.8 & 68.6 &53.0 &47.2 \\
\rowcolor{lightgray} GCDG (ours) & \bfseries69.3           & \bfseries24.7 & 55.5        & \bfseries17.1 & 68.9&\bfseries55.5 & \bfseries48.5 \\
    \bottomrule
    \end{tabular}
    }
    \vspace{-3mm}
\end{table}

\subsection{Experimental Results on Classification}

Table~\ref{PACS} presents the out-of-domain performances on PACS, showcasing that GCDG outperforms existing SOTA methods on average accuracy. Notably, GCDG achieves the highest accuracy on the hard-to-transfer domains, namely `Cartoon' and `Sketch'. This success can be attributed to the generative classifier, which is able to capture the valuable domain-specific information that aids classification in those hard-to-transfer domains. The suboptimal performance of GCDG in `photo' can be attributed to the saturated performance level in `photo'.

Table~\ref{terra} provides the results on Terra-Incognita~(TI), reaffirming GCDG's superiority as it outperforms SOTA methods in hard-to-transfer domains and achieves the best average accuracy. This consistency further underscores GCDG's merit of accommodating diverse features.

As shown in Table~\ref{office_home}, the highest average accuracy on Office-Home~(OH) with diverse backbones further emphasizes the superiority of the proposed GCDG, although Office-Home presents a more challenging benchmark due to its larger number of categories.

The performance on VLCS is summarized in Table~\ref{VLCS}. While GCDG may not achieve the best performance in individual scenarios, its highest average accuracy showcases its efficacy in maintaining robust performance across various scenarios instead of excelling in a specific one.

We report the generalization performance on DomainNet~(DN) in Table~\ref{domainnet}. DomainNet contains a larger dataset with images from 345 classes, presenting a great challenge for DG. Despite this challenge, our proposed GCDG achieves the best generalization in at least three out of six scenarios. Besides, GCDG showcases the highest average accuracy across diverse domains. These findings demonstrate the superiority of the proposed GCDG in boosting model generalizability.

\subsection{OCIM Face Anti-spoofing}
To further demonstrate the effectiveness and versatility of our proposed GCDG, we additionally conduct experiments on a different computer vision task, \emph{i.e.}, Face anti-spoofing (FAS). FAS aims to enhance the robustness of models in distinguishing between real and spoofed faces, thereby protecting the face recognition system from various attacks. Following common protocols in DG-FAS~\cite{jia2020single, liu2021adaptive, liu2021dual, shao2019multi, shao2020regularized, zhou2022adaptive}, we conduct experiments on OCIM benchmark comprising four diverse datasets: CASIAMFSD~\cite{zhang2012face} (C), Idiap Replay-Attack~\cite{chingovska2012effectiveness} (I), and MSU-MFSD~\cite{wen2015face} (M), OULUNPU\cite{boulkenafet2017oulu} (O), and report the leave-one-out-validation performance on Half Total Error Rate~(HTER) and Area Under Curve~(AUC).

\noindent
\textbf{Implementational Details.} To ensure a fair comparison, we utilize the same network architecture as~\cite{liu2021adaptive, liu2021dual, shao2020regularized, zhou2023instance}, and extract images in RGB channels with the input size as $256\times256\times3$. For optimization, we choose Adam for the backbone with a learning rate of 0.0005. Following previous research~\cite {liu2021adaptive, liu2021dual, zhou2023instance}, PRNet~\cite{feng2018joint} is utilized to attain pseudo-depth signals for depth supervision.

\noindent
\textbf{Comparison Results on Leave-One-Out settings.} 
Table~\ref{fas} presents the generalization of state-of-the-art methods and our proposed approach on FAS.
Notably, GCDG achieves the best HTER performance across all testing scenarios among the SOTA methods.
Regarding AUC, GCDG attains the highest average performance across diverse testing settings, demonstrating its effectiveness in mitigating the adverse effects of distribution shifts in FAS.
These results underscore the superiority of GCDG in enhancing generalization by modeling diverse domain-specific information rather than merely learning decision boundaries.

\section{Empirical Analysis}
\label{section_5}
\begin{table}[]
    \centering
    \caption{Generalization performance with various classifiers on five commonly used DG datasets. MLP-ERM denotes ERM with MLP-based classifier.}
    \setlength{\tabcolsep}{3pt}
    \resizebox{0.49\textwidth}{!}{%
    \begin{tabular}{c|ccccc|c}
    \toprule
     Model & PACS ($\uparrow$) & VLCS ($\uparrow$) & OH ($\uparrow$) & TI ($\uparrow$) & DN ($\uparrow$) & Avg. ($\uparrow$) \\
     \midrule
     ERM                     & 80.72 & 74.50 & 60.51 & 41.44 & 43.68 & 60.17\\
     MLP-ERM                     & 81.38 & 73.85 & 61.24 & 43.17 & 42.68 & 60.47\\
     \midrule
     GCDG (ours)                            & \bfseries83.44 & \bfseries76.86 &\bfseries64.49 & \bfseries44.86 &\bfseries46.63 & \bfseries63.26\\
     \bottomrule
    \end{tabular}
    }
    \label{MLP}
    \vspace{-3mm}
\end{table} 

In this section, we conduct an in-depth analysis to elucidate the proposed GCDG's superiority and gain insight into its underlying mechanisms.

\begin{figure*}
    \centering
    \includegraphics[width=0.95\textwidth]{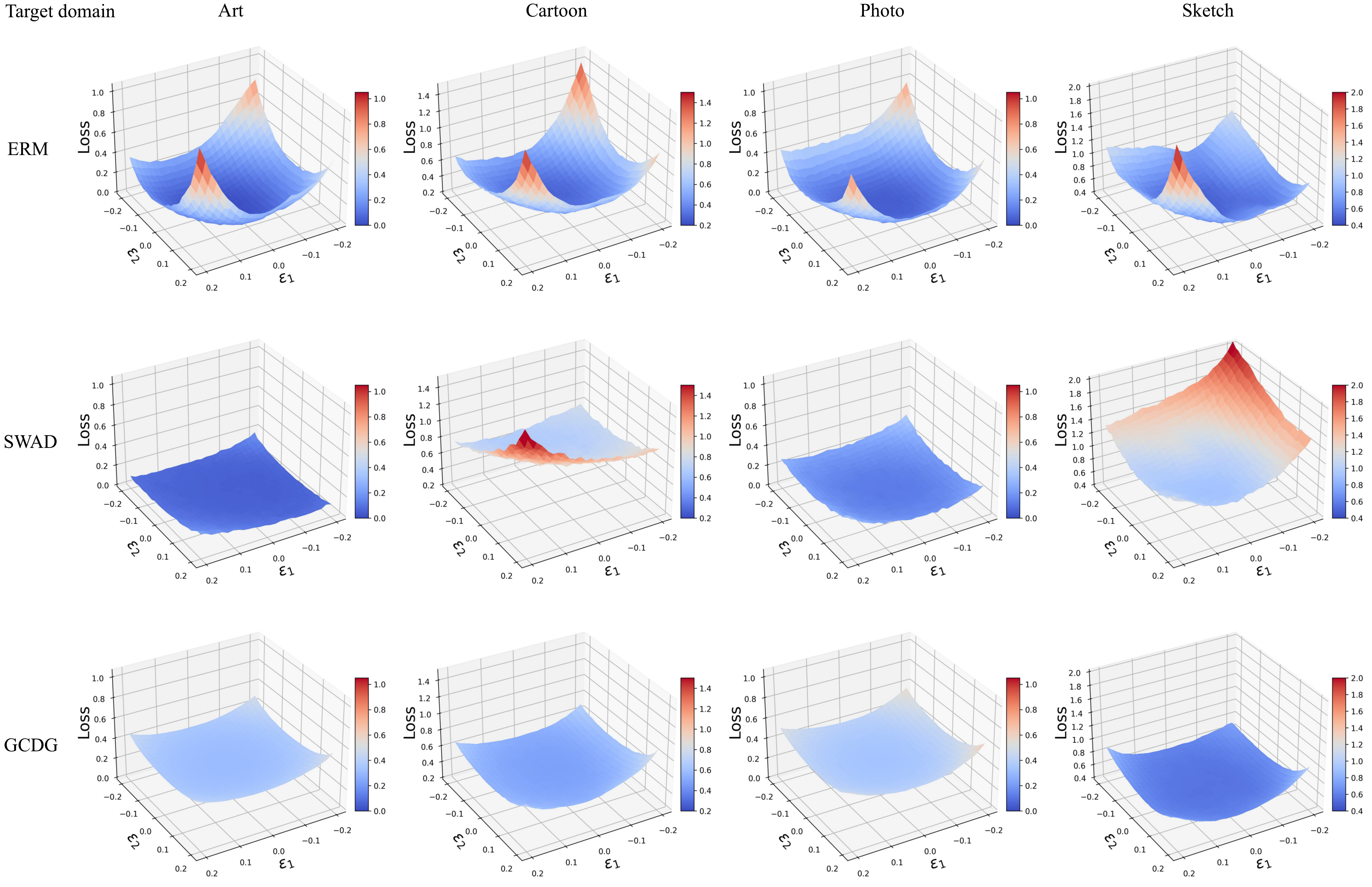}
    \caption{Visualization of the loss landscapes for ERM, the flatness-aware method SWAD~\cite{cha2021swad}, and the proposed GCDG on PACS. Note that the loss landscape is visualized on the source domains. Notably, our proposed GCDG exhibits superior efficacy in fostering flat minima compared to ERM and the flatness-aware method SWAD.}
    \label{visual_swad}
    \vspace{-3mm}
\end{figure*}

\noindent\textbf{Neural Network-based Classifier \emph{vs.} Generative Classifier.}
To demonstrate the efficacy of the proposed GCDG model in capturing intricate multi-modal distributions, we conducted an empirical comparison with an alternative approach: over-parameterizing the classifier by employing a Multi-Layer Perceptron (MLP) as a substitute for the linear classifier. The MLP is implemented with two fully-connected layers, featuring an equivalent or larger parameter count in comparison to the proposed generative classifier. It is noteworthy that non-linear activations are incorporated. The out-of-domain generalization results, presented in Table~\ref{MLP}, underscore the superior capability of the GCDG model in representing a diverse array of data patterns. The observed advantage of the generative classifier over the MLP classifier can be attributed to the fact that the MLP primarily focuses on discerning the decision boundary between different classes rather than comprehensively capturing the inherent patterns and structures within the data, as achieved by the generative classifier. Consequently, the generative classifier exhibits superior performance when confronted with new samples from the target domains.

\begin{table}[t]
    \centering
    \setlength{\tabcolsep}{2pt}
    \caption{Ablation study of the proposed components in GCDG on PACS.}
    \resizebox{0.48\textwidth}{!}{
    \begin{tabular}{c|ccc|cccc|c}
    \toprule

    \multirow{2}{*}{ID} & \multirow{2}{*}{HLC} & \multirow{2}{*}{DCB} & \multirow{2}{*}{SCB} & \multicolumn{4}{c|}{Target domain} & \multirow{2}{*}{Avg.($\uparrow$)}\\
    \cmidrule(r){5-8}
    &&&& Art & Cartoon & Photo & Sketch & \\
    \midrule
    A                & -          & -           & -  &78.76&74.79&96.29& 73.02   & 80.72 \\
     B      & \checkmark        & -         & -   &81.84&74.49&94.13&  77.35   & 81.95\\
    C       & \checkmark        & \checkmark         & -   &81.74&74.27&93.71& 79.82    & 82.39\\
    D       & \checkmark        &  -        & \checkmark   &82.13&75.85&94.91& 76.28    & 82.29\\
     E     & \checkmark        & \checkmark         & \checkmark  &83.06&78.50&92.63&  79.56    & 83.44\\
    \bottomrule
    \end{tabular}} 
    \label{ablation}
    \vspace{-5mm}
\end{table}

\noindent\textbf{Ablation Study.}
To demonstrate the efficacy of the proposed components in our GCDG, we conduct ablation study on PACS to illustrate their contributions. As shown in Table~\ref{ablation}, compared to ERM (model A) where there are no DG techniques, our proposed HLC could remarkably enhance the generalization performance, resulting from its capability of leveraging valuable domain-specific information. Besides, DCB prevents the unbalanced contributions of components in GMMs, further boosting the model generalizability.
Futhermore, the designed SCB could help HLC avoid capturing spurious correlations, thereby unleashing the potential of our model to take advantage of essential domain-specific information.

\noindent\textbf{Loss Landscape Visualization.} To visually illustrate how the proposed GCDG results in flat minima in the loss landscape, we provide quantitative results to visualize the loss landscapes. Following the local loss landscape visualization in~\cite{li2018visualizing}, we plot the loss landscapes on source domains by choosing two direction vectors and perturbing them. As observed in Fig.~\ref{visual_swad}, the loss landscapes by incorporating GCDG demonstrate significant improvement in flatness compared to ERM in all scenarios. Furthermore, GCDG exhibits superior performance in the pursuit of flat minima over SWAD~\cite{cha2021swad}, a strategy known for its effectiveness in seeking flat minima. These findings are consistent with the claim that GCDG could promote flat minima for better generalization performance.

\noindent\textbf{Leveraging Domain-specific Information.}
Our proposed GCDG introduces the generative classifier to unlock the potential of valuable domain-specific information, thereby reducing the source risk. 
To demonstrate the superiority of GCDG in leveraging domain-specific information, we compare it with existing methods that utilize domain-specific information, namely DMG~\cite{chattopadhyay2020learning}, DDG\cite{zhang2022towards}, and DRM\cite{zhang2023domain}. 
Specifically, DMG learns domain-specific masks to encourage a balance of domain-invariant and domain-specific features.
DDG employs representation disentanglement to learn feature variations, which are then used to generate novel samples. 
DRM introduces a test-time adaptation strategy that dynamically ensembles domain-specific classifiers based on information from unseen test samples. For a fair comparison, we constrain DRM to ensemble models based on uniform weights rather than the dynamic weights calculated using test samples.
Table~\ref{DDG&DRM} reports the generalization performance. The 2.8\% improvement over DRM highlights the superiority of our GCDG in fully leveraging valuable domain-specific information, which is achieved through modeling the underlying feature distributions via the generative classifier.

\begin{table}[]
    \centering
    \caption{Comparison between methods leveraging domain-specific information.}
    \setlength{\tabcolsep}{4pt}
    \resizebox{0.49\textwidth}{!}{%
    \begin{tabular}{l|cccc|c}
    \toprule
        \multirow{2}{*}{Models}& \multicolumn{4}{c|}{Target domain}  & \multirow{2}{*}{Avg.($\uparrow$) } \\
        \cline{2-5}
        & Art & Cartoon & Photo & Sketch \\
         \toprule
         DMG \cite{chattopadhyay2020learning} & 76.90 &  \bfseries80.38& 93.35 & 75.21&  81.46\\
         \midrule
          DDG \cite{zhang2022principled} & 79.30 & 74.00& 91.80 & 75.80&  80.20\\
         \midrule
          DRM \cite{zhang2023domain} &81.20 & 71.20 & \bfseries93.70 & 78.60 & 81.20 \\
         \midrule
         GCDG (ours) & \bfseries83.06 &78.50& 92.63 & \bfseries79.56&  \bfseries83.44 \\
         \bottomrule
    \end{tabular}}
    \label{DDG&DRM}
    \vspace{-5mm}
\end{table}

\noindent\textbf{Plug-and-Play with other DG Methods.} Our GCDG adopts a simple yet effective approach by replacing the prevalent linear classifier in DG with a generative classifier, aiming to model the diverse distributions across domains, which is a challenging task for linear classifiers. As such, GCDG is orthogonal to existing DG approaches, allowing for straightforward integration with existing methods by merely substituting the linear classifier, with no need for changes to the feature extractor or training procedures. Here, we integrate GCDG into ERM, SWAD~\cite{cha2021swad}, and PCL~\cite{yao2022pcl}, without fine-tuning the hyperparameters of the generative classifier. The results presented in Table~\ref{conbination} demonstrate that GCDG consistently enhances the performance of DG models and exhibits new SOTA performance when combined with PCL.

\noindent\textbf{Computational Efficiency.}
To evaluate the computational efficiency of the proposed generative classifier in GCDG, we compare its computational cost against discriminative classifiers, including the linear probe (ERM) and the non-linear discriminative classifier (MLP-ERM).
Here, MLP-ERM refers to the ERM algorithm with a non-linear MLP classifier, designed with more parameters than GCDG.
Comparison metrics include model parameters, floating-point operations per second (FLOPs), and inference time.
The results are presented in Table~\ref{flops}.
As observed, GCDG achieves improved generalization performance with negligible computational overhead compared to ERM.
Furthermore, the increased number of parameters in GCDG effectively enhances generalizability, as the generative classifier captures feature distributions rather than merely learning decision boundaries, as in the discriminative classifier of MLP-ERM.

\begin{table}[]
    \centering
    \caption{Integration of the proposed GCDG into DG methods on PACS.}
    \setlength{\tabcolsep}{4pt}
    \resizebox{0.49\textwidth}{!}{%
    \begin{tabular}{l|cccc|c}
    \toprule
        \multirow{2}{*}{Models}& \multicolumn{4}{c|}{Target domain}  & \multirow{2}{*}{Avg.($\uparrow$) } \\
        \cline{2-5}
        & Art & Cartoon & Photo & Sketch \\
         \midrule
          ERM & 78.76 & 74.79& \bfseries96.29 & 73.02&  80.72\\
        \ + GCDG & \bfseries83.06 & \bfseries78.50& 92.63 & \bfseries79.56&  \bfseries83.44\\
         \midrule
          SWAD \cite{cha2021swad} & 83.28 & 74.63 & \bfseries96.56 & 77.96 & 83.11  \\
         \ + GCDG &\bfseries85.17 & \bfseries78.04& 95.14 & \bfseries78.28 & \bfseries84.16 \\
         \midrule
        PCL \cite{yao2022pcl} & 83.53 & 73.61& \bfseries96.18 & 77.20&  82.63\\
        \ + GCDG & \bfseries83.83 & \bfseries78.36& 95.66 & \bfseries80.31&  \bfseries84.54 \\
         \bottomrule
    \end{tabular}}
    \label{conbination}
\end{table}

\begin{table}[]
    \centering
    \caption{Comparison of computational efficiency. MLP-ERM denotes ERM with MLP-based classifier. Tested with the image size of $224\times224$ on one NVIDIA Tesla V100 GPU.}
    {
    \begin{tabular}{c|c|c|c}
    \toprule
     Model & \# of Params (M) & GFlops  & Time (ms) \\
     \midrule
    ERM           & 11.180         & 1.82167 & 19.727\\
    MLP-ERM           & 11.212         & 1.82170 & 20.183\\
    GCDG           & 11.188         & 1.82167 & 20.764\\
     \bottomrule
    \end{tabular}}
    \label{flops}
    \vspace{-5mm}
\end{table}

\section{Conclusion}

In this work, we present a generative paradigm for DG classifier, which aims to address the drawbacks associated with the mainstream domain-invariant methods and the prevalent linear classifier. 
Through theoretical analysis, we underscore the necessity of incorporating domain-specific information for better generalization performance. Building upon this fact, we highlight the shortcomings of the commonly used linear classifier in capturing valuable domain-specific information exhibiting multi-modality. 
To effectively leverage the crucial domain-specific information and solve the limitations inherent in the linear classifier, we propose a novel method, named GCDG, to replace the linear classifier with a generative classifier. 
GCDG comprises three key modules: Heterogeneity Learning Classifier (HLC), Spurious Correlation Blocking (SCB), and Diverse Component Balancing (DCB). HLC models the multimodal data distributions to effectively leverage domain-specific information. SCB mitigates the adverse effects of spurious correlations on HLC. Furthermore, DCB ensures balanced contributions of components within HLC.
These advantages empower our proposed approach to diminish the upper bound of target risk and promote flat minima. 
The proposed GCDG shows superior performance compared to existing DG methods on five DG benchmarks and one FAS benchmark.
As a versatile plug-and-play module for DG, GCDG can be seamlessly integrated with other approaches to enhance generalization capacity. 
We believe that this work could open up a novel direction for DG, and inspire more future works that leverage the full potential of domain-specific information via a generative framework.

\section*{Acknowledgment}
The authors would like to thank Dr. Xiangtai Li, Prof. Yunhai Tong, and Prof. Dacheng Tao for their insightful suggestions and stimulating discussions.

\ifCLASSOPTIONcaptionsoff
  \newpage
\fi

\bibliographystyle{IEEEtran}
\bibliography{dg}

\clearpage
\appendices

\section{Proof}

\subsection{Proof of Theorem 1.}

\begin{proof}
Given the celebrated data-processing inequality~\cite{beaudry2011intuitive, cover1999elements}, for 1 $\leq $ $i$ $\leq $ $M$, it follows that:
\begin{equation}
    I(Z_i; Y) \leq I(X_i; Y).
\end{equation}
On the other hand, if $p(Z_1, Y) = \cdots = p(Z_i, Y) = \cdots = p(Z_M, Y)$, the following holds:
\begin{equation}
    I(Z_1; Y) = \cdots = I(Z_i; Y) = \cdots = I(Z_M; Y).
\end{equation}
In accordance with the variational form of conditional entropy~\cite{farnia2016minimax,zhao2022fundamental}, we obtain:
\begin{equation}
    \begin{split}
    &\inf_{g} \sum_{i = 1}^{M}\mathbb{E}_{p_i}[\ell_{CE}(g(Z_i), Y)] - \inf_{h} \sum_{i = 1}^{M}\mathbb{E}_{p_i}[\ell_{CE}(h(X_i), Y)] \\
    = & \sum_{i = 1}^{M} H(Y|Z_i) - \sum_{i = 1}^{M} H(Y|X_i) 
    =  \sum_{i = 1}^{M} I(X_i; Y) - \sum_{i = 1}^{M} I(Z_i; Y) \\
    \geq & \sum_{i = 1}^{M} I(X_i; Y) - \sum_{i = 1}^{M} \max I(Z_i; Y) \\
    \geq & \sum_{i = 1}^{M} I(X_i; Y) - M \min_{1\leq i \leq M}\{ I(X_i; Y)\} \\
    \geq & \sum_{i \neq m^*} (I(X_i; Y) - I(X_{m^*}; Y)) = \Delta_p,
    \end{split}
\end{equation}
where $\Delta_p$ is the information gap of source domains defined in the main text.
\end{proof}

\subsection{Proof of Theorem 2.}

\begin{proof}
We express the constrained infimums for both hypothesis spaces in terms of the unconstrained Bayes risks $\epsilon_T^*(Z)$ and $\epsilon_T^*(Q)$ plus their respective non-negative approximation errors $\mathcal{A}()$.
For features $Z$:
\begin{equation}
    \inf_{h \in \mathcal{H}_Z} \epsilon_T(h) = \inf_{f \in \mathcal{F}_{all}} \epsilon_T(f) + \mathcal{A}(\mathcal{H}_Z) = \epsilon_T^*(Z) + \mathcal{A}(\mathcal{H}_Z).
\end{equation}
For features $Q$:
\begin{equation}
    \inf_{h \in \mathcal{H}_Q} \epsilon_T(h) = \inf_{f \in \mathcal{F}_{all}} \epsilon_T(f) + \mathcal{A}(\mathcal{H}_Q) = \epsilon_T^*(Q) + \mathcal{A}(\mathcal{H}_Q).
\end{equation}

For cross-entropy loss, the Bayes-optimal risk equals the conditional entropy of the target label given the representation space:
$$\epsilon_T^*(Z) = H_T(Y \mid Z),   \epsilon_T^*(Q) = H_T(Y \mid Q).$$
Subtracting $\epsilon_T^*(Q)$ from $\epsilon_T^*(Z)$ yields:
\begin{equation}
    \epsilon_T^*(Z) - \epsilon_T^*(Q) = H_T(Y \mid Z) - H_T(Y \mid Q).
\end{equation}
Consider the chain rule for conditional entropy across $Y \to Q \to Z$, the following inequality holds:
$$H_T(Y \mid Z) - H_T(Y \mid Q) = I_T(Q; Y \mid Z) := \Delta_Q \ge 0.$$
Therefore, the exact Bayes risk difference is:$$\epsilon_T^*(Z) - \epsilon_T^*(Q) = \Delta_Q$$

Take the difference between the constrained infimums:
\begin{equation}
    \inf_{h \in \mathcal{H}_Z} \epsilon_T(h) - \inf_{h \in \mathcal{H}_Q} \epsilon_T(h) = \left( \epsilon_T^*(Z) + \mathcal{A}(\mathcal{H}_Z) \right) - \left( \epsilon_T^*(Q) + \mathcal{A}(\mathcal{H}_Q) \right).
\end{equation}
Rearrange terms:
\begin{equation}
    \inf_{h \in \mathcal{H}_Z} \epsilon_T(h) - \inf_{h \in \mathcal{H}_Q} \epsilon_T(h) = \underbrace{\left( \epsilon_T^*(Z) - \epsilon_T^*(Q) \right)}_{= \Delta_Q} +\underbrace{\left( \mathcal{A}(\mathcal{H}_Z) - \mathcal{A}(\mathcal{H}_Q) \right)}_{\text{Approximation Gap Difference}}.
\end{equation}

Because $Z$ is a compressed transformation of $Q$ ($Q \to Z$), representation $Z$ drops feature dimensions and structural variance present in $Q$. Fitting a constrained parametric function class $\mathcal{G}_Z$ on a compressed space $\mathcal{Z}$ is at best as effective as, and typically worse than, fitting $\mathcal{G}_Q$ on the richer space $\mathcal{Q}$. Thus we obtain:
\begin{equation}
    \mathcal{A}(\mathcal{H}_Z) \ge \mathcal{A}(\mathcal{H}_Q) \implies \mathcal{A}(\mathcal{H}_Z) - \mathcal{A}(\mathcal{H}_Q) \ge 0,
\end{equation}
substituting this into the infimums difference gives:
$$
    \inf_{h \in \mathcal{H}_Z} \epsilon_T(h) - \inf_{h \in \mathcal{H}_Q} \epsilon_T(h) \ge \Delta_Q + 0
$$
\begin{equation}
     \implies \inf_{h \in \mathcal{H}_Z} \epsilon_T(h) - \inf_{h \in \mathcal{H}_Q} \epsilon_T(h) \ge \Delta_Q,
\end{equation}
where $\Delta_Q = I_T(Q; Y \mid Z) \ge 0$.
\end{proof}

\clearpage

\clearpage

\end{document}